\documentclass[9pt,journal]{IEEEtran}
\usepackage{comment}
\usepackage{graphicx}
\usepackage{tikz}
    \usetikzlibrary{shapes.misc,shapes.geometric,arrows.meta,positioning,shadows.blur,chains,scopes}
    \usetikzlibrary{positioning,calc}
    \tikzset{
    block1/.style = {rectangle, fill=yellow!16, draw, align=center, minimum height=2.5cm},
    blockb/.style = {rectangle, fill=blue!16, draw, align=center, minimum height=2.5cm},
    blockg/.style = {rectangle, fill=green!16, draw, align=center, minimum height=2.5cm}
    }
    
\usepackage{cleveref}

\usepackage{hyperref} 
\hypersetup{
    colorlinks,
    citecolor=black,
    filecolor=black,
    linkcolor=black,
    urlcolor=black
}

\usepackage{colortbl} 

\ifCLASSINFOpdf
\else
\fi
\usepackage{mathtools}

\DeclarePairedDelimiter\abs{\lvert}{\rvert}

\ifCLASSOPTIONcompsoc
 \usepackage[caption=false,font=normalsize,labelfont=sf,textfont=sf]{subfig}
\else
 \usepackage[caption=false,font=footnotesize]{subfig}
\fi
\hypersetup{draft}
\begin{document}
%
\title{Transfer Learning for Non-Intrusive Load Monitoring}
%
%
%

\author{Michele~D'Incecco, 
        Stefano~Squartini, ~\IEEEmembership{Senior Member,~IEEE, CIS Member,~IEEE,}
        and~Mingjun~Zhong
\thanks{M. D'Incecco was with Dipartimento di Ingegneria dell'Informazione - Università Politecnica delle Marche, Via Brecce Bianche 12, 60131 Ancona, Italy. E-mail:micheledincecco@yahoo.it.}
\thanks{S. Squartini was with Dipartimento di Ingegneria dell'Informazione - Università Politecnica delle Marche, Via Brecce Bianche 12, 60131 Ancona, Italy. E-mail: s.squartini@univpm.it.}
\thanks{M. Zhong was with the School of Computer Science in the University of Lincoln, Lincoln LN6 7TS, UK. E-mail: mzhong@lincoln.ac.uk.}}
\maketitle

\begin{abstract}
Non-intrusive load monitoring (NILM) is a technique to recover source appliances from only the recorded mains in a household. NILM is  unidentifiable and thus a challenge problem because the inferred power value of an appliance given only the mains could not be unique. To mitigate the unidentifiable problem, various methods incorporating domain knowledge into NILM have been proposed and shown effective experimentally. Recently, among these methods, deep neural networks are shown performing best. Arguably, the recently proposed sequence-to-point (seq2point) learning is promising for NILM. 
However, the results were only carried out on the same data domain. It is not clear if the method could be generalised or transferred to different domains, e.g., the test data were drawn from a different country comparing to the training data. We address this issue in the paper, and two transfer learning schemes are proposed, i.e., appliance transfer learning (ATL) and cross-domain transfer learning (CTL). For ATL, our results show that the latent features learnt by a `complex' appliance, e.g., washing machine, can be transferred to a `simple' appliance, e.g., kettle. For CTL, our conclusion is that the seq2point learning is transferable. Precisely, when the training and test data are in a similar domain, seq2point learning can be directly applied to the test data without fine tuning; when the training and test data are in different domains, seq2point learning needs fine tuning before applying to the test data. Interestingly, we show that only the fully connected layers need fine tuning for transfer learning. Source code can be found at https://github.com/MingjunZhong/transferNILM.

\end{abstract}


%
\IEEEpeerreviewmaketitle

\section{Introduction}
%
%
%
%

\IEEEPARstart{A}{s} the climate changes, the governments are committed to reduce CO$_2$ emissions, which triggers the demand to reduce energy consumption globally \cite{archer2012global,ro06900t}. This requires to understand how the energy was used and thus lead to the optimised energy management which could finally help us to consume energy efficiently. Therefore energy management has become nowadays an increasing research field, which particularly is drawn attention to by machine learning community due to the emerging large scale data which needs to be understood in principle. 

In household energy management, appliance load monitoring is a research topic which investigates to retrieve the real-time energy consumption of each appliance within household buildings \cite{s121216838,7165334}. Since the purpose is to monitor all the appliances in a household, the instant power readings need to be recorded. Perhaps the easiest way to obtain those power readings is to deploy smart sensor devices for each appliance which are able to record the power consumption. 
These data could be used to retrieve the entire information about the energy consumption of the household buildings. However, this method is an intrusive load monitoring approach, and in addition each device is expensive to install and hard to maintain at the same time. In order for avoiding the use of sensor devices, an alternative approach which is called Non-Intrusive Load Monitoring (NILM) \cite{hart1992} was proposed to recover the energy consumption of each appliance. The aim of NILM is to extract energy consumption of each appliance in a household by only using the mains readings.

    When NILM is applied to domestic buildings, it can inform the end-users the power consumption and consequently it could help householders to reduce the energy consumption. Studies have shown that, a feedback to the users is a useful tool to aid the users to understand how the energy was used, which acts as a ``\textit{self-teaching tool}" \cite{1168}. The feedback to the users provided by smart meters could positively affect the energy consumption behaviours. Research has shown that NILM could help household users to reduce energy consumption by 15\% \cite{fischer2008feedback}. It is even more useful for the energy providers which can optimise their smart grid operations on one side and propose specific tariffs for users according to their energy consumption habits.
    
    Recently, machine learning methods \textemdash \hspace{0.1cm} typically deep learning algorithms \textemdash\hspace{0.1cm} have been greatly developed with the help of big data. Specifically in energy management, this encourages researchers to develop machine learning approaches to NILM since energy data of hundreds of households have been and are being collected from different countries. Two kinds of machine learning algorithms, i.e., unsupervised and supervised learning, are the main methods for NILM. Among those algorithms, deep learning approaches have achieved the state-of-the-art performance. Particularly, it has been shown in literature that convolutional neural networks (CNN) are able to extract meaningful latent features for appliances which are particularly useful for NILM and thus achieves the best performance \cite{s2p}. In the original seq2point learning paper \cite{s2p}, the algorithm was trained and tested on the same domain, that is, it was trained on UK-DALE and tested on UK-DALE, and trained on REDD and tested on REDD. Our motivation in this paper is that it was not clear if the seq2point learning is transferable and what could be the strategies for transfer capability. An oracle NILM would be to train the models over a number of houses and appliances and then deploy for prediction on any other unseen houses and unseen appliances. For example, we expect that NILM is trained over UK houses and deployed in US, and trained over washing machine and deployed for kettle. In this paper, we propose two transfer learning strategies for tackling these challenges approaching to the oracle NILM. Our hypothesis is that those features extracted by using CNN are invariant across appliances and as well as across data domains. These invariant features are the key factors to develop reliable transfer learning algorithms targeting the oracle NILM system. To investigate this hypothesis, we propose appliance transfer learning (ATL) and cross-domain transfer learning (CTL). For ATL, the target is to evaluate the algorithms if the features learnt by one appliance are able to be transferred to other appliances. For CTL, the target is to evaluate the algorithms if they could be transferred from one data domain to another. Our experiments have validated both of the transfer learning strategies which could potentially lead to the ultimate NILM. Our strategies can greatly reduce the number of training samples from unseen houses and unseen appliances. Specifically, transfer learning is interesting in terms of training a NILM because of the following reasons: 1) The ground truth active power data for each appliance is expensive to obtain. Transfer learning could potentially reduce the number of sensors for each appliance to be installed for  obtaining  training  data since available models could be transferred to other appliances or domains. 2) Transfer learning also offers remarkable computational savings as pre-trained models can be reused for other appliances or domains. 
     
    \par The structure of this paper is as follows: In section \ref{Background and literature review} a background of NILM is presented; Section \ref{Methodology} presents the seq2point learning algorithm and as well as the transfer learning; Experimental results are shown in sections \ref{Experiments} and \ref{expresults}, and finally we draw our conclusions.

\section{Background}\label{Background and literature review}

This section will introduce the NILM model in details. The existing machine learning approaches to NILM will be reviewed. These approaches are mainly classified into unsupervised and supervised learning methods. 

    Firstly we summarise the problem of non-intrusive load monitoring. Suppose that the mains $Y(t)$ are the aggregation of all the active power consumption of the individual appliances in a household. Note that $Y(t)$ denotes the mains reading at time $t$. Then the mains could be represented as the following formula,
        \begin{equation}
            Y(t)=  \sum\limits_{i=1}^{I} X_i(t) + e(t),
            \label{eq:pot}
        \end{equation}
    where $X_i(t)$ represents the power reading of the appliance $i$ at time $t$, $I$ is the number of appliances and $e(t)$ is the variable representing the model noise. It is quite often that the noise variable follows a Gaussian with mean $0$ and variance $\sigma_t$, i.e., $e(t)\sim \mathcal{N}(0,\sigma^2(t))$. Given the observed mains $Y=\{Y(t)\}_{t=1}^T$, our task is then to recover all the appliances $X_i=\{X_i(t)\}_{t=1}^T (i=1,\cdots,I)$. Clearly, this is a Single-channel Blind Source Separation problem, which means that given only a single observation we want to recover more than one sources. It is well-known  that this problem is non-identifiable. Various approaches have been proposed to approach the NILM by alleviating the identifiability issue. Naturally, it is believed that this problem could be alleviated if the domain knowledge could be incorporated into the model. In literature, various methods have been proposed for tackling NILM, for example, signal processing approaches which use features of appliances for the purpose of disaggregations \cite{WANG2018134,berges2011user,en12060992,en11123409,jin2011time}, clustering algorithms \cite{barsim2016sequentialB,liu2019}, matrix factorization \cite{7049328,batra2018transferring}, etc. There are mainly two approaches to NILM using machine learning, which are unsupervised and supervised learning. We now briefly review the existing algorithms belonging to these two approaches.
    
    \subsubsection{Unsupervised learning}
    The dominated approach in unsupervised learning is the additive factorial hidden Markov model (AFHMM) \cite{kolter2012approximate,zhong2014signal,Parson12,BONFIGLI20171590}. This is a natural approach to modelling the NILM problem if we admit the following hypothesis: 1) an appliance is a realisation of a hidden Markov model; 2) the power reading of the mains at a time is the addition of the appliances at that time. Under these assumptions, NILM can be represented as an AFHMM. The aim is then to infer the most probable states of those appliances. However, the problem is still unidentifiable. 
    
    In order for improving the performance of AFHMM, constraints were imposed on the model from different aspects. The main approach was to constrain the model with domain knowledge. For example, local information such as appliance power levels, ON-OFF state changes and duration can be incorporated into the model \cite{kolter2012approximate,Parson12,pattem2012unsupervised,Zhao15,Tabatabaei17}. The methods in \cite{kolter2012approximate,zhong13} proposed to use change-one-at-a-time constraint on the appliances. Global information such as total energy consumption, total number of cycles, and the total time of an appliance being occupied was also proposed to incorporate into the model  \cite{zhong2014signal,zhong2015latent,Batra16,Stephen17}. Signal processing methods were also proposed to incorporate useful information into NILM \cite{pattem2012unsupervised,Zhao15,BOUHOURAS2019392,He7539273}. However, the main limitation of these approaches is that the domain knowledge needs to be extracted manually from the observation data, which makes the method difficult to use.
    
    \subsubsection{Supervised learning}
    Recently, many household electricity data have been published. These data have both the mains and correspondingly the appliance power readings, which makes it possible to formulate NILM as a supervised learning. Precisely, we have observed the pairs of the data $(X_t,Y_t)$ where $X_t$ and $Y_t$ denote respectively the power reading of an appliance and the mains at time $t$. Because there are plenty of observations, it is possible to train a model to represent the relationship between $X$ and $Y$. It is thus reasonable to learn a function $f$ to represent this relationship so that
    \begin{equation}
         X = f(Y),
    \end{equation}
    which can be viewed as a non-linear regression problem.
    Deep neural networks is a natural approach to learn the function $f$, and recently, it has been successfully applied to NILM \cite{kelly2015neural,s2p,shin2018subtask,chen2018convolutional,bao2018enhancing,barsim2018feasibility,bonfigli2018denoising,murray2018transferability,de2018appliance,brewitt2018non,ebrahim2018energy,kim2017nonintrusive,Stephen19}. From the point of the view of statistical methods, it has been pointed out in \cite{s2p} that given the pairs $(X,Y)$ we could train a model to represent a conditional distribution $p(X|Y)$. Recently, both sequence-to-sequence and sequence-to-point learning methods were proposed to approximate this distribution \cite{s2p}. Interestingly, those desired features, for example, appliance power levels, ON-OFF state changes and duration, were able to be learnt by the neural networks. These features do not need to be extracted manually.
    
    As NILM is proposed as a non-linear regression problem, the function $f$ could be represented as various kinds of forms. In recent literature, various architectures of the neural networks representing $f$ have been proposed, which are de-noising auto-encoder \cite{kelly2015neural,bonfigli2018denoising}, generative adversarial networks \cite{bao2018}, convolutional neural networks (CNN) \cite{s2p,shin2018subtask,chen2018convolutional,murray2018transferability,dash2018very}, recurrent neural networks (RNN) \cite{kelly2015neural}. Among these methods, CNN was believed performing the best \cite{s2p,shin2018subtask}. However, the hyper-parameters of the architectures, e.g, number of layers, could affect the performance of those neural networks. 
    
    It should be noted that unknown  appliances, base  load, and  as  well  as  model  noise are typically important for certain unsupervised learning approaches, e.g., FHMMs. Due to the nature of this class of models, these variables have to be explicitly modelled. However, deep learning approaches do not have to  model  these  variables explicitly.  The  goal  of  deep neural networks (DNN)  is  to  extract  the  target,  i.e.,  the  appliance,  and  treat all the other variables as background. In other words, our DNN-based approach for NILM could be viewed as a denoising technique, which retrieves the specific appliance power consumption from the aggregate one, therefore the  noise  contribution  are  automatically  removed  together  with  the  consumption  contributions  of other concurrent appliances.

\section{Methods}\label{Methodology}

    In this section, the methodologies employed in this paper are described. Most of the deep learning methods were assessed on the same domain. Less work has been done on the generalizability except the work of \cite{murray2018transferability}, in which the authors considered cross-domain transfer learning. However, their work does not study fine tuning and appliance transfer learning. Sparse coding approach was proposed for transferring knowledge across regions \cite{batra2018transferring}, however, it would be a challenge to scale to large data as it involves optimising large matrix factorization. In this paper we investigate transferability (generalizability) of deep neural networks applying to NILM in wider aspects. Among neural based techniques for NILM, the sequence-to-point learning is very well suited for transfer learning strategies, due to its architecture characteristics. We will thus consider the transferability of the seq2point learning. Two transfer learning approaches are considered which are appliance transfer learning and cross-domain transfer learning. Firstly, the seq2point learning is presented as follows.
    \subsection{Sequence-to-point learning}
\begin{figure*}
            \centering
            \includegraphics[width=\linewidth,height=0.5\textheight,keepaspectratio]{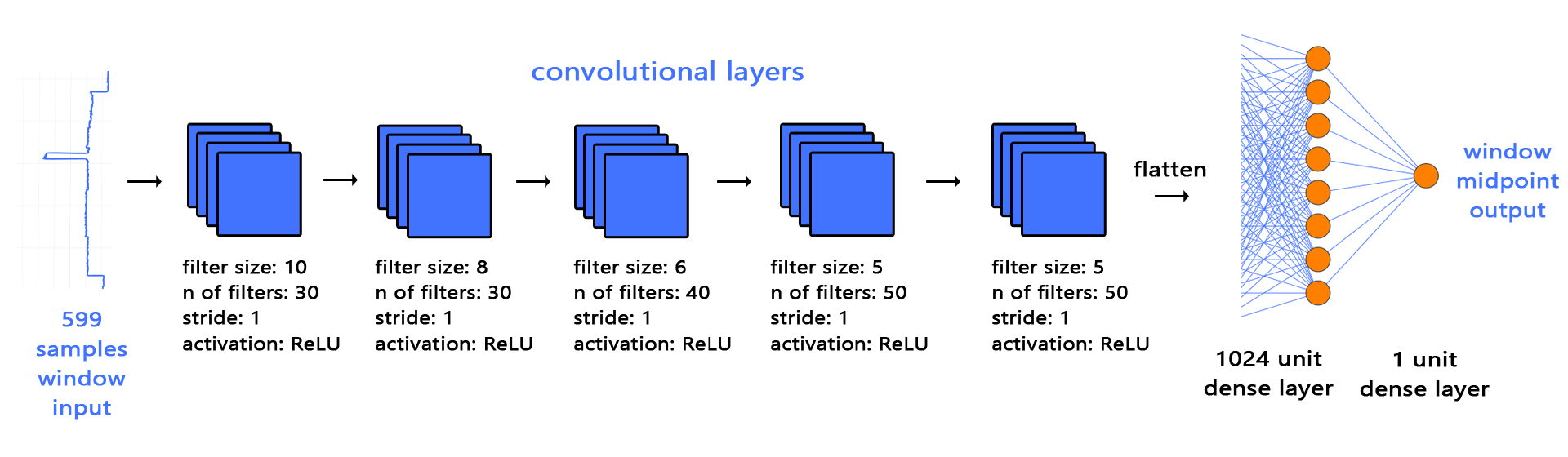}
            \caption{Architecture for the sequence-to-point learning.}
            \label{fig:s2p}
        \end{figure*}

    The main idea of the seq2point learning is to learning a neural network to represent the midpoint of an appliance given a window of the mains being the input. See the Figure \ref{fig:s2p} for the architecture for seq2point learning. Precisely, the input of the network is a mains window $Y_{t:t+W-1}$, and the output is the midpoint element $x_{\tau}$ of the corresponding window of the target appliance,
    where $\tau=t+W/2$.
    The representation assumes that the midpoint element is a non-linear function of the mains window. The intuition behind this assumption is that we expect the state of the midpoint element of that appliance to relate to the information of mains before and after that midpoint.
    
    Instead of mapping sequence to sequence, the seq2point architectures
define a neural network $F_p$ which maps sliding windows $Y_{t:t+W-1}$ of
the input to the midpoint $x_{\tau}$ of the corresponding windows
$X_{t:t+W-1}$ of the output. The model is
$x_{\tau}=f(Y_{t:t+W-1})+\epsilon$. The loss function used
for training has the 
form 
\begin{equation}
\label{seq2point}
L_p = \sum_{t=1}^{T-W+1} \log p(x_{\tau} | Y_{t:t+W-1}, \theta),
\end{equation}
where $\theta$ are the network parameters. 
 To deal with
the endpoints of the sequence, given a full 
input sequence $Y = (y_1 \ldots y_T)$, we first pad the sequence
with $W/2$ 
zeros at the beginning and end. 
The advantage of the seq2point model is that 
there is a single prediction for every $x_t$, rather than
an average of predictions for each window.

It has been shown that seq2point learning performs well on UK-DALE and REDD data \cite{s2p}. For UK-DALE, the method was trained on the houses 1,3,4 and 5 and tested on the house 2; for REDD, the method was trained on houses $2-6$ and tested on house 1. This indicates that the seq2point learning was only evaluated on the same domain. However, the ability of generalisation is unclear. To deploy a system employing seq2point method, it is crucial to assess the generalisability of the method. In the following sections, we consider appliance transfer learning and cross-domain transfer learning.

    \subsection{Appliance transfer learning}\label{Sequence to point transfer learning}
\begin{figure*}            \includegraphics[width=1\linewidth,height=1\textheight,keepaspectratio]{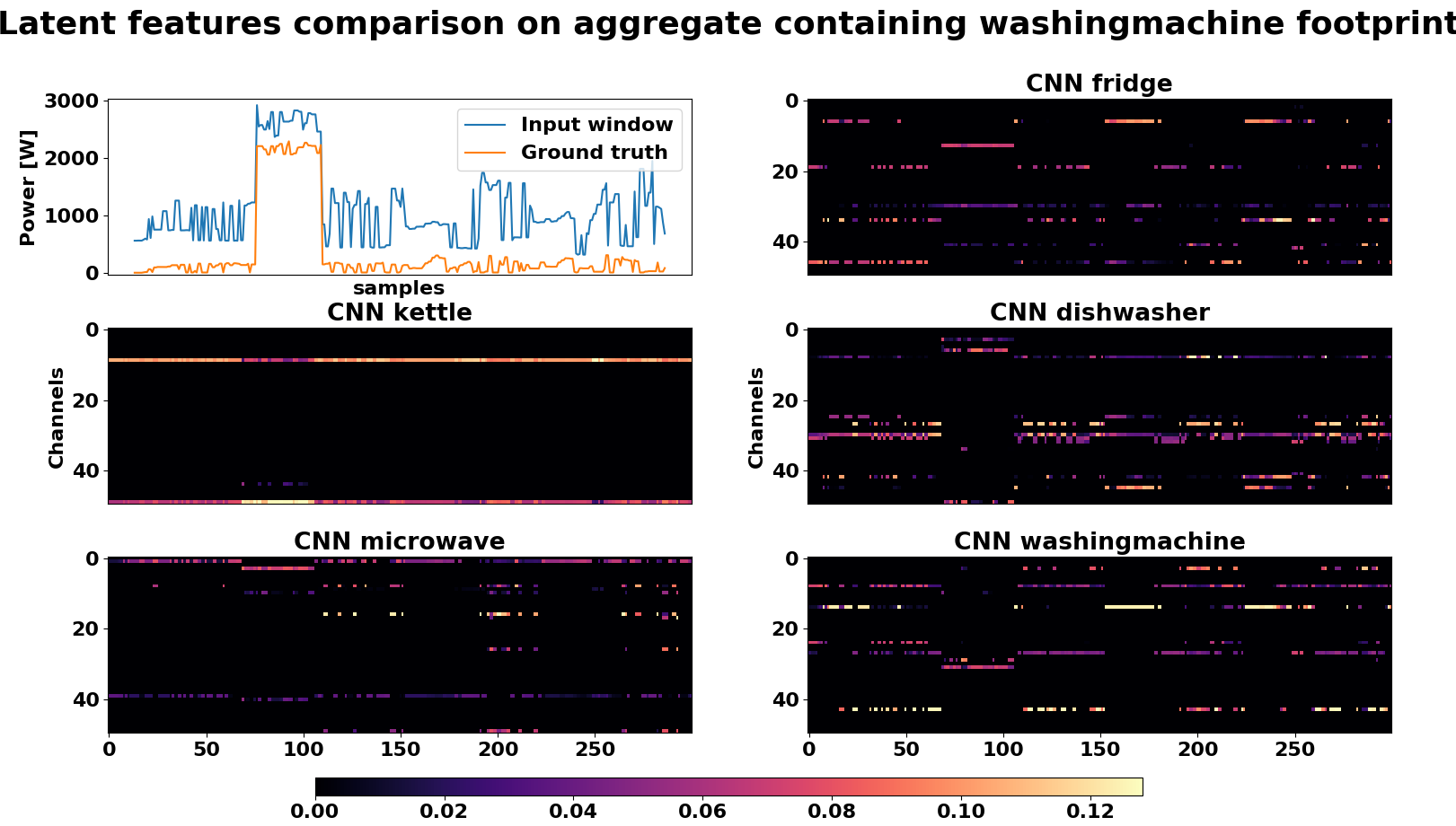}
             \centering
             \caption{The features learnt by using seq2point learning. The first row shows the input to the networks and the ground truth of an appliance. The other rows show the outputs of the last convolutional layer of seq2point models trained by using kettle, microwave, fridge, dish washer, and washing machine.}
             \label{fig:features}
         \end{figure*}

    We consider appliance transfer learning (ATL). Our question is that if the signatures learnt by one appliance could be transferred to other appliances. We observed that the features learnt by difference appliances have the similar patterns. It is interesting to note that mainly the features were ON-OFF changes, the power levels, and the duration of an activity. For example, Figure \ref{fig:features} demonstrates the features produced by the models trained by using kettle, microwave, fridge, dish washer and washing machine data. In this figure, all the five models used the same input window. Our observation is that the features learnt by different appliances are often different, although they are learning similar signatures. Some appliances, e.g., washing machine and dish washer, extract more information than others, e.g., kettle. Most of the feature channels from the kettle were not active; for example, in Figure \ref{fig:features}, only two channels actively have significant signatures. Comparing to kettle, the washing machine learned more information; many more channels are actively showing interesting signatures.
    
    Since all the appliances are learning similar signatures from the data, it is possible to share the same feature channels across all the appliances. If this was feasible, potentially this approach could greatly help to reduce the training costs and the number of sensors to be installed in a household for obtaining training data. Our approach is then to train a model using only an individual appliance. The trained CNN layers will be directly applied to other appliances, and only the fully connected layers will be trained for a different appliance. We choose to use the CNN layers trained by using washing machines.
    
        \begin{figure}
             
             \includegraphics[width=\linewidth,height=\textheight,keepaspectratio]{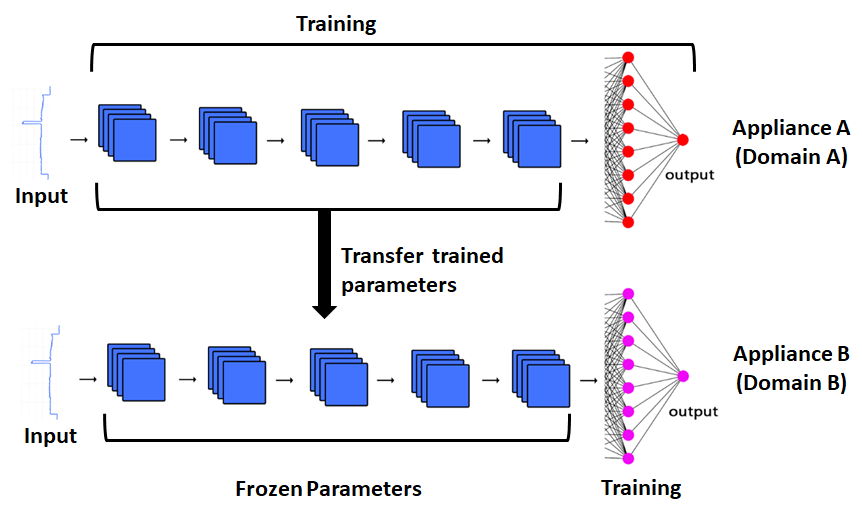}
             \centering
             \caption{Transfer learning used in this paper. For the appliance transfer learning, the CNN layers trained by appliance A are transferred to appliance B and frozen; only the fully connected layers are trained for appliance B. For cross-domain transfer learning, if domain A and domain B are similar, both CNN and fully connected layers trained in domain A are transferred to B; if they are different, the CNN layers are transferred from A to B and only fully connected layers are trained on domain B.}
             \label{fig:transferlearning}
         \end{figure}

         \subsection{Cross domain transfer learning}\label{Transfer learning}
         We are interested in cross domain transfer learning (CTL) for seq2point learning. In this paper, the different domain means that the data were collected from different regions either in the same or a different country. The model can be trained on one domain and tested on a different domain. Our aim is to evaluate the model and investigate how to transfer the model from one domain to another. Our purpose is to transfer the model by using as little information as possible from the test domain. We will employ fine tuning for transfer learning.
         
         The seq2point model is trained against a large data set. Then, two transfer learning approaches will be investigated in this paper. The first approach is to directly apply the trained model to a test data in a different domain; the second approach is to fine tune the pre-trained model by using a small subset from the test data.  It is interesting to note that we will not tune the CNN layers for both approaches, because our experiments suggested that tuning CNN layers do not improve the performance. Therefore, only the full connected layers are tuned in our transfer learning methods. Both ATL and CTL are illustrated in Figure \ref{fig:transferlearning}.

\section{Experimental Setup}\label{Experiments}
    \subsection{Data sets}
    Several open-source data sets are available for the purpose of energy disaggregation. These data were measured in household buildings from different countries. The sensors installed in these buildings read active power, but some sensors also read other information, for example, reactive power, current, and voltage. In NILM, the active power data are used. However, the main difference between the data sets is the sampling frequency. Due to this issue, pre-processing for aligning the readings need to be done before NILM algorithms are applied to the data.  In literature, five appliances are usually considered for disaggregation \cite{s2p,kelly2015neural}, which are kettle, microwave, fridge, dish washer and washing machine. In our experiments, three household electricity data sets will be used, which are REFIT \cite{refit}, UK-DALE \cite{ukdale}, and REDD \cite{redd}.
    
        \subsubsection{REFIT}
         The REFIT data were collected from 20 buildings in England (Loughborough area) and cover a period between 2013 and 2015 \cite{refit}. The data contain active power sampled every 8 seconds for both mains and individual appliances.
    
        This data set is the largest among the three data sets in our study, and therefore is used for training the deep learning models. We expected that the large amount of electricity readings would be able to generalise the trained deep learning methods, which can then be applied to other unseen houses. Note that before training the models, the whole data were inspected and we found that houses 13 and 21 were using solar panel energy production, which were thus not used for our current research.
        
        \subsubsection{UK-DALE}
        The UK-DALE (UK Domestic Appliance-Level Electricity) contains 5 buildings in the UK during the period from 2013 to 2015. The sampling periods for mains and appliances were $1s$ and $6s$, respectively. Further details and statistics can be found in \cite{ukdale}.
        
        \subsubsection{REDD}
        The Reference Energy Disaggregation Data Set (REDD) is a data measured for 6 buildings in US. Measurements include mains with $1s$ sampling period and several appliances with $3s$ sampling period. High-frequency current and voltage measurements are also available at $15KHz$ sample frequency. The lengths of observations were between 3 and 19 days. 

        \subsubsection{Preparing training and test data}
        The deep learning models for appliances were trained individually, which indicates that the model of an appliance was trained by only using the data of that appliance. When preparing the training data, we firstly inspect the mains and appliance readings by visualising them. It is very often that both the mains and appliances were missing large chunks of readings, which may be caused when the hardware went wrong. For example, the sensors may be turned off or the batteries ran out. These large chunks of missing data were then removed from the training data. In other cases, some missing data were small, for example, the missing data were just from seconds to minutes. These missing data were kept in the data because they could be treated as noise, which could lead to a better regularized model. The splits for the REFIT data are shown in the Table \ref{tab:tabREF}, which were used for all of our experiments. 
        \begin{table*}[htbp]
\caption{Distribution of the REFIT dataset. M: Millions; Y: years.}
                \centering
                \scalebox{1}{
                \begin{tabular}{|l|l|l|l|l|l|l|l|l|l|}
                \hline
                                     & \multicolumn{3}{|c|}{Training} & \multicolumn{3}{|c|}{Validation}        & \multicolumn{3}{|c|}{Test}\\
                \hline
                Appliance                    &
                \multicolumn{1}{|c|}{house} & 
                \multicolumn{1}{|c|}{samples (M)} &
                \multicolumn{1}{|c|}{time (Y)}&
                \multicolumn{1}{|c|}{house} & 
                \multicolumn{1}{|c|}{samples (M)} &
                \multicolumn{1}{|c|}{time (Y)}&
                \multicolumn{1}{|c|}{house}  &
                \multicolumn{1}{|c|}{samples (M)} &
                \multicolumn{1}{|c|}{time (Y)}\\
                
                \hline
                Kettle      &3, 4, 6, 7, 8, 9 &59.19 &15  &5 &7.43  &1.9  &2  &5.73 &1.5\\
                          &12, 13, 19, 20  &&&&&&&&\\
                \hline
                Microwave           &10, 12, 19 &18.22 &4.6     &17 &5.43 &1.4     &4 &6.76 &1.7\\
                \hline
                Fridge              & 2, 5, 9 &19.33 &4.9     &12 &5.86 &1.5      &15 &6.23 &1.6\\
                \hline
                Dish w.          & 5, 7, 9, 13, 16   &30.82 &9.8    &18  &5 &1.3      & 20 &5.17 &1.3\\
                \hline
                Washing m.     & 2, 5, 7, 9, 15, 16, 17   &43.47 &11 &18 &5 &1.3  &8 &6 &1.5\\
                \hline
                \end{tabular}
                }
                
                \label{tab:tabREF}
            \end{table*}
        
The buildings 1 and 2 from UK-DALE were used for our experiments since the data in other buildings are small. All the five appliances from buildings 1 and 2 were used for testing. There were 4.01 millions data samples recorded in around 12 months which were used for our experiments. Since sample period in UK-DALE was $6s$, comparing to the REFIT which has $8s$, the UK-DALE data were down-sampled to $8s$.

In REDD data, four appliances which are microwave, fridge, dish washer and washing machine from house 1, 2, and 3 were used for our experiments. Each appliance has around 1.2 million data samples which were recorded in about 4 months. It should be noted that kettle was not recorded in this data set.

    All the data need to be preprocessed before they were used for training and testing. Firstly, the data were normalized by using the following formula,
    \begin{equation}
        \frac{x_t-\bar{x}}{\sigma},
    \end{equation}
    where $x_t$ denotes a reading at time $t$, $\bar{x}$ denotes the mean value of an appliance or the mains, and $\sigma$ denotes the standard deviation of an appliance or the mains. The mean and standard deviation values used for normalisation are shown in the Table \ref{tab:norm}. These values were only used for the purpose of normalising the data, instead of informing the means and variances of those appliances. After the data were normalised, they can then be fed into the models for training.
    
            \begin{table}[htbp]
            \centering
            \caption{Parameters for normalising the data}
            \begin{tabular}{|l|c|c|}
            \hline
            & mean & standard deviation\\
            \hline
            Aggregate           & 522     & 814 \\
            \hline
            Kettle             & 700    & 1000\\
            \hline
            Microwave           & 500     & 800\\
            \hline
            Fridge              & 200    & 400 \\
            \hline
            Dishwasher          &  700   & 1000 \\
            \hline
            Washing machine     &  400   & 700\\
            \hline
            \end{tabular}
             \label{tab:norm}
             
            \vspace{0.5cm}
           
            \caption{The parameters used for training the models}
            \centering
            \begin{tabular}{|l|l|}
            \multicolumn{1}{l}{\textbf{Hyper-parameters for training}}\\
            \hline
            Input window size (samples)                       & 599 \\
            \hline
            Number of maximum epochs              & 50               \\
            \hline
            Batch size                & 1000  \\
            \hline
            Minimum early-stopping epochs   & 5     \\
            \hline
            Patience of early-stopping (in epochs)  & 5     \\
            \hline
            
            \multicolumn{1}{l}{\textbf{Parameters used for ADAM optimiser}}\\
            \hline
            Learning rate                & 0.001    \\
            \hline
            Beta1                & 0.9    \\
            \hline
            Beta2               & 0.999    \\
            \hline
            Epsilon             & $10^{-8}$ \\
            \hline
            \end{tabular}
            \label{tab:AdamHP}
        \end{table}
        
    \subsection{Metrics}\label{Metrics}
    Four metrics will be employed in this paper to evaluate the algorithms. The first metric is the mean absolute error (MAE), which evaluates the absolute difference between the prediction $\hat{x}_t$ and the ground truth $x_t$ at every time point and calculates the mean value
        
        \begin{equation}
            MAE = \frac{1}{T} \sum\limits_{t=1}^{T} \abs{\hat{x}_t - x_t}.
        \end{equation}
    The second metric is the normalised signal aggregate error (SAE), which indicates the relative error of the total energy. Denote $r$ as the total energy consumption of the appliance and $\hat{r}$ the predicted total energy, then SAE is defined as
        
        \begin{equation}
                SAE = \frac{\abs{\hat{r} - r}}{r}.
        \end{equation}
    The third metric is the energy per day (EpD): 
        
        \begin{equation}
            EpD = \frac{1}{D} \sum\limits_{n=1}^{D} \abs{\hat{e} - e},
        \end{equation}
    where $e=\sum\limits_{t} x_t $ denotes the energy consumed in a day period and $D$ is the total number of days. This metric indicates the absolute error of the predicted energy used in a day, which is typically useful when the household users are interested in the total energy consumed in a period. The fourth metric is the normalized disaggregation error (NDE):
    \begin{equation}
        NDE = \frac{\sum_{i,t}(x_{it}-\hat{x}_{it})^2}{\sum_{it}x^2_{it}}
    \end{equation}
    This metric measures the normalized error of the squared difference between the prediction and the ground truth of the appliances. 
    
    \subsection{Settings for training neural networks}
    In accordance with the network's architecture of the sequence-to-point learning, a fixed-length window of aggregate active power consumption signal is given as input. For the neural networks employed in this paper, a sample window has 599 data points. However, other window lengths should also be evaluated. The best length could be selected experimentally, for example the work in \cite{shin2018subtask} studied the best window length for sequence-to-sequence model. In our experiments, a sample window was generated by sliding the window forward by a single data point, and so all the possible sample windows were used for training. The windows of the mains were used as the inputs to the neural networks, whilst the midpoints of the corresponding windows of the appliances were used as the targets. 
    
\begin{figure}
                \centering
                \includegraphics[width=\linewidth,height=1\textheight,keepaspectratio]{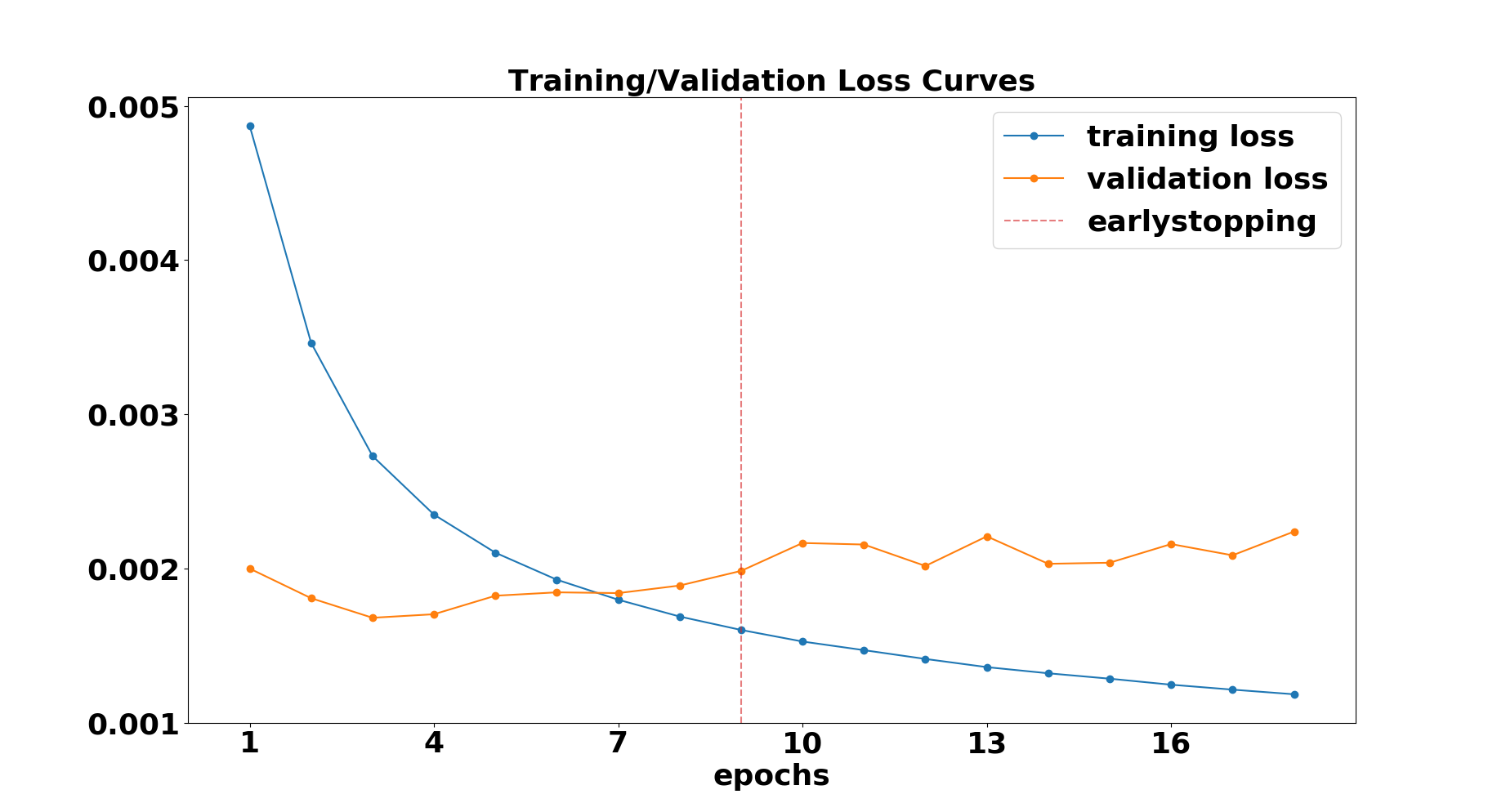}
                \caption{The training and validation losses. The vertical dashed line indicates that the iteration stopped at the $9^{th}$ epoch when the patience of early stopping was set to 6.}
                \label{fig:loss}
            \end{figure}

    Tensorflow were used to train the model. The ADAM optimiser algorithm \cite{adam} was used for training and the early-stopping criterion was employed to reduce overfitting. The hyper-parameters for training and as well as the ADAM optimizer parameters are shown in the Table \ref{tab:AdamHP}. Note that dropout were not used in our experiments, which could be explored in future development. It is known that neural networks could be overfitting when training too long. We used early-stopping to decide when to stopping training. Figure \ref{fig:loss} shows how we applied early-stopping. We monitored both the losses for training and validation data and used 6 iterations as our patience for early-stopping. It can be seen that the algorithm achieved minimum validation loss at the $3^{rd}$ epoch which was used as our final trained model. 

\begin{table*}[h]
\caption{Appliance transfer learning for sequence-to-point learning. W.M. denotes Washing Machine.}
                \centering
                \scalebox{1}{
                \begin{tabular}{|l|c|c|c|c|c|c|c|c|c|c|c|c|c|c|c|}
                \hline
                       & \multicolumn{4}{|>{}c|}
                        {AFHMM} & \multicolumn{4}{|>{}c|}
                        {Trained on REFIT}&  \multicolumn{4}{|>{}c|}{CNN trained on REFIT (using W.M.)}\\
                         
                        & \multicolumn{4}{|>{}c|}
                        {} &  \multicolumn{4}{|>{}c|}{Tested on REFIT}&  \multicolumn{4}{|>{}c|}{Tested on REFIT}\\
                \hline
                                     &\multicolumn{1}{|c|}{MAE}&\multicolumn{1}{|c|}{SAE}& \multicolumn{1}{|c|}{EpD [Wh]}&\multicolumn{1}{|c|}{NDE}&\multicolumn{1}{|c|}{MAE}&\multicolumn{1}{|c|}{SAE}& \multicolumn{1}{|c|}{EpD [Wh]}&\multicolumn{1}{|c|}{NDE}&\multicolumn{1}{|c|}{MAE} & \multicolumn{1}{|c|}{SAE}&\multicolumn{1}{|c|}{EpD [Wh]}&\multicolumn{1}{|c|}{NDE}\\
                \hline
                Kettle      &74.37 &2.41 &1203.52 &2.17    &6.830    &0.130   &153.92& 0.52  &12.690    &0.050   &96.04 &0.20 \\
                \hline
                Microwave   &12.25 &0.84 &276.90 &0.99   &12.660    &0.170   &95.78 &0.71   &13.380    &0.050   &107.730&0.70 \\
                \hline
                Fridge   &61.12 &0.99 &1305.55 &0.99       &20.020    &0.330   &216.85 & 0.43   &20.560    &0.440   &270.48 &0.45  \\
                \hline
                Dish w.   &163.79 &1.10 &3238.71 &5.12  &12.260   &0.260   &180.00 &0.44  &12.360   &0.610   &206.15 & 0.49\\
                \hline
                Washing m. &48.09 &0.80 &866.02 &1.47 &16.850    &2.610   &319.11 &0.54   &16.850    &2.610   &319.11 & 0.54\\
                \hline
                Overall   &71.92 &2.88  &1378.14 &2.15  & 13.724    & 0.702      & 193.13 &0.52 & 15.168    &0.752      &149.62 &0.47   \\
                mean $\pm$ std &50.38 &3.68 &997.24 &1.55  & 4.477    & 0.957      &74.32 &0.11  & 3.135    &0.955      &87.765 &0.18  \\
                \hline
                \end{tabular}
                }
                
                \label{tab:tabrestransF}
            \end{table*}
        
        
    \section{Experimental Results}
    \label{expresults}
    The sequence-to-point learning were original applied to relatively small data sets which are UK-DALE and REDD \cite{s2p}. The model was only evaluated at the same domains, which means that the model was trained on UK-DALE and then tested on the UK-DALE data, and the same procedure was applied to the REDD data. However, it is unclear if the seq2point learning could be generalised in different domains. It is well known that deep neural networks are often overfitting on a domain. This often occurs when the training domain has a significantly different distribution to the testing domain. In NILM application, the household data were collected from different countries. The appliances may have significantly different patterns in different countries. Even in the same country, the patterns of the appliances may also be different from different households. Our experiments will investigate the transferability of seq2point learning and show how the seq2point learning could be used for transfer learning.
    
    Firstly, the seq2point learning was trained and test in the same domain. 
    Tables \ref{tab:tabrestransF}, \ref{tab:tabukdale_comp}, and \ref{tab:rreddcomp} show the results on training and testing on REFIT, UK-DALE and REDD respectively. Note that the results on REDD and UK-DALE were taken from \cite{s2p}. As a baseline, the AFHMM method \cite{zhong2014signal} was also applied to these data and the results are shown in the first columns in Tables \ref{tab:tabrestransF}, \ref{tab:tabukdale_comp}, and \ref{tab:rreddcomp}.
    
    Secondly, seq2point was trained on REFIT and tested on UK-DALE and REDD. The third columns in Tables \ref{tab:tabukdale_comp} and \ref{tab:rreddcomp} show the results on UK-DALE and REDD respectively. In terms of MAE and SAE, testing on UK-DALE the transfer learning interestingly improves the performance of seq2point learning; however, testing on REDD, the transfer learning did not improve the performance of seq2pint learning. The reason could be that REFIT and UK-DALE are in the similar domain because the data were collected in the same country. REFIT is a much larger data than UK-DALE, and so the model was better generalised. REDD does not have the similar domain to REFIT, which might be the reason that transfer learning was not improving the performance of seq2point learning.
  
    \subsubsection{Appliance transfer learning}         According to the analysis in \ref{Sequence to point transfer learning}, the CNN trained on washing machine will be used to all the other appliances. The CNN layers were trained using REFIT. In the appliance transfer learning, those trained CNN layers will be used directly as the CNN layers for other appliances. Only the fully-connected layers are trained for other appliances such as kettle, microwave, fridge and dishwasher. Note that the training hyper parameters in table \ref{tab:AdamHP} were used for training the dense layers. Since the CNN layers were not trained, the training time was reduced to a half of the time for training all the network parameters. 

\begin{figure}
                \centering
                \includegraphics[width=\linewidth,height=1\textheight,keepaspectratio]{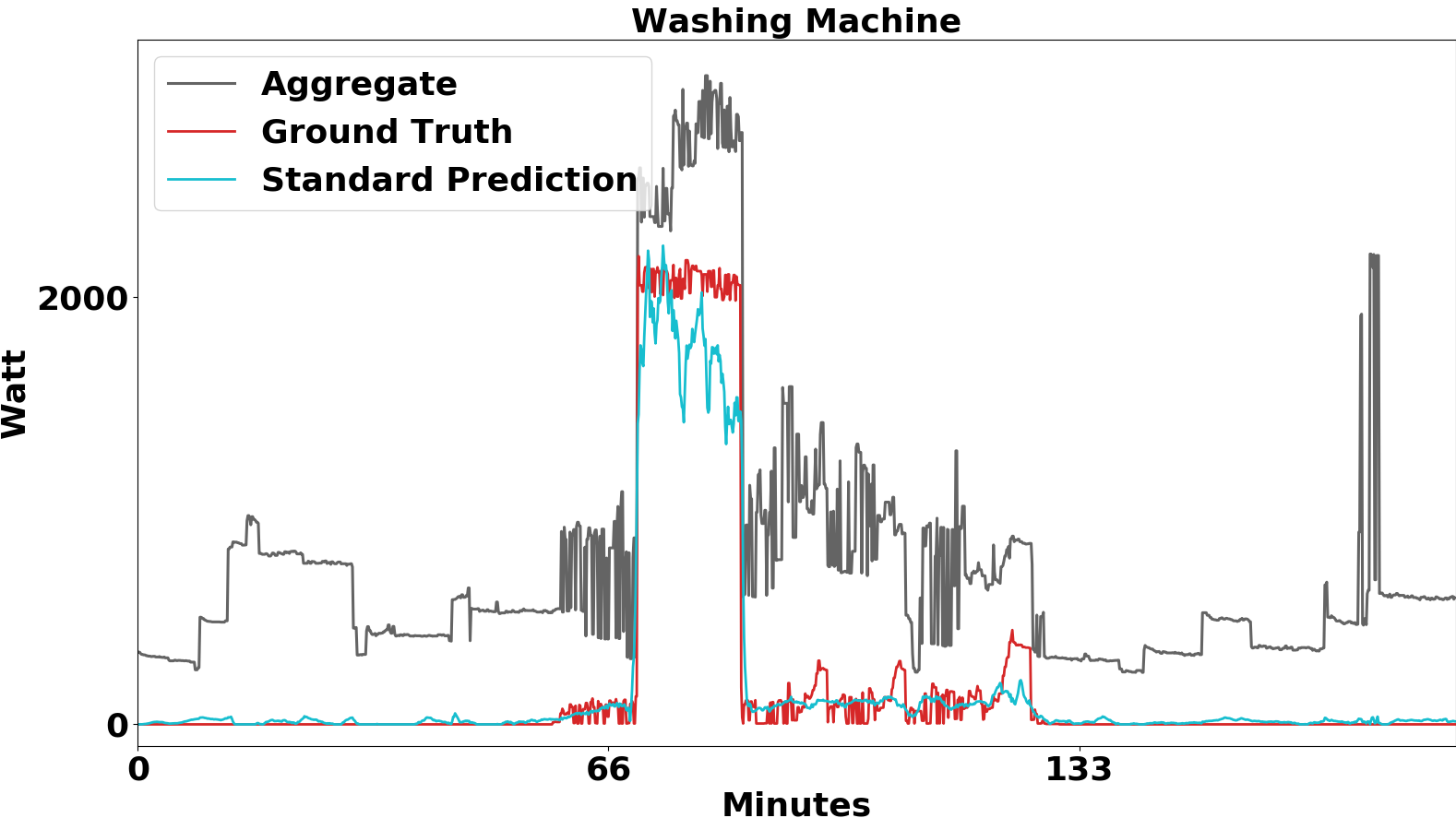}
                \caption{Prediction for washing machine using seq2point learning.}
                \label{fig:wash}
            \end{figure}

\begin{figure}
                \centering
                \includegraphics[width=\linewidth,height=1\textheight,keepaspectratio]{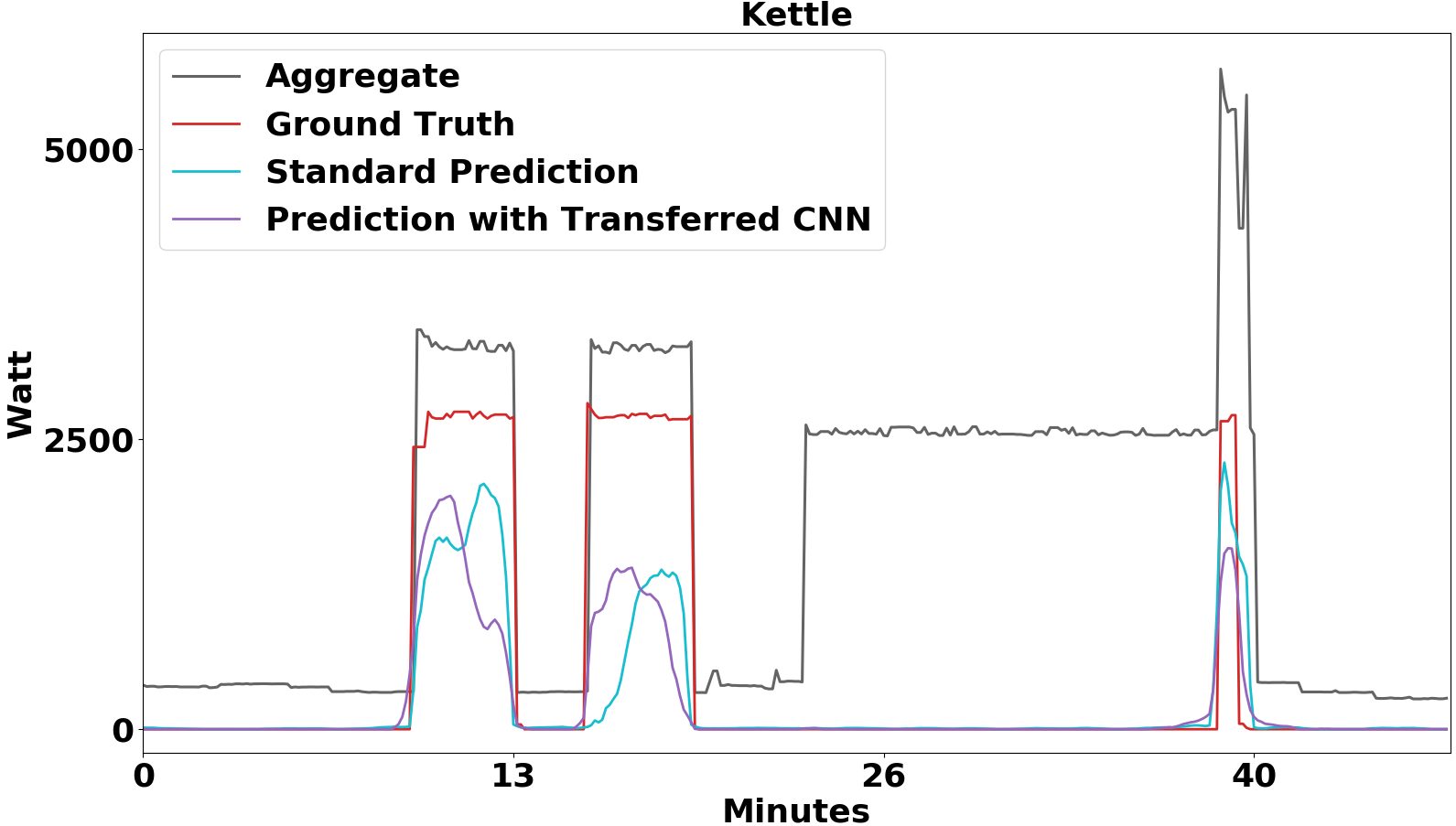}
                \caption{Standard prediction using seq2point model trained on kettle data and the prediction using appliance transfer learning where the CNNs were trained on washing machine data.}
                \label{fig:pred-comp-kettle}
            \end{figure}

\begin{figure}
                \centering
                \includegraphics[width=\linewidth,height=1\textheight,keepaspectratio]{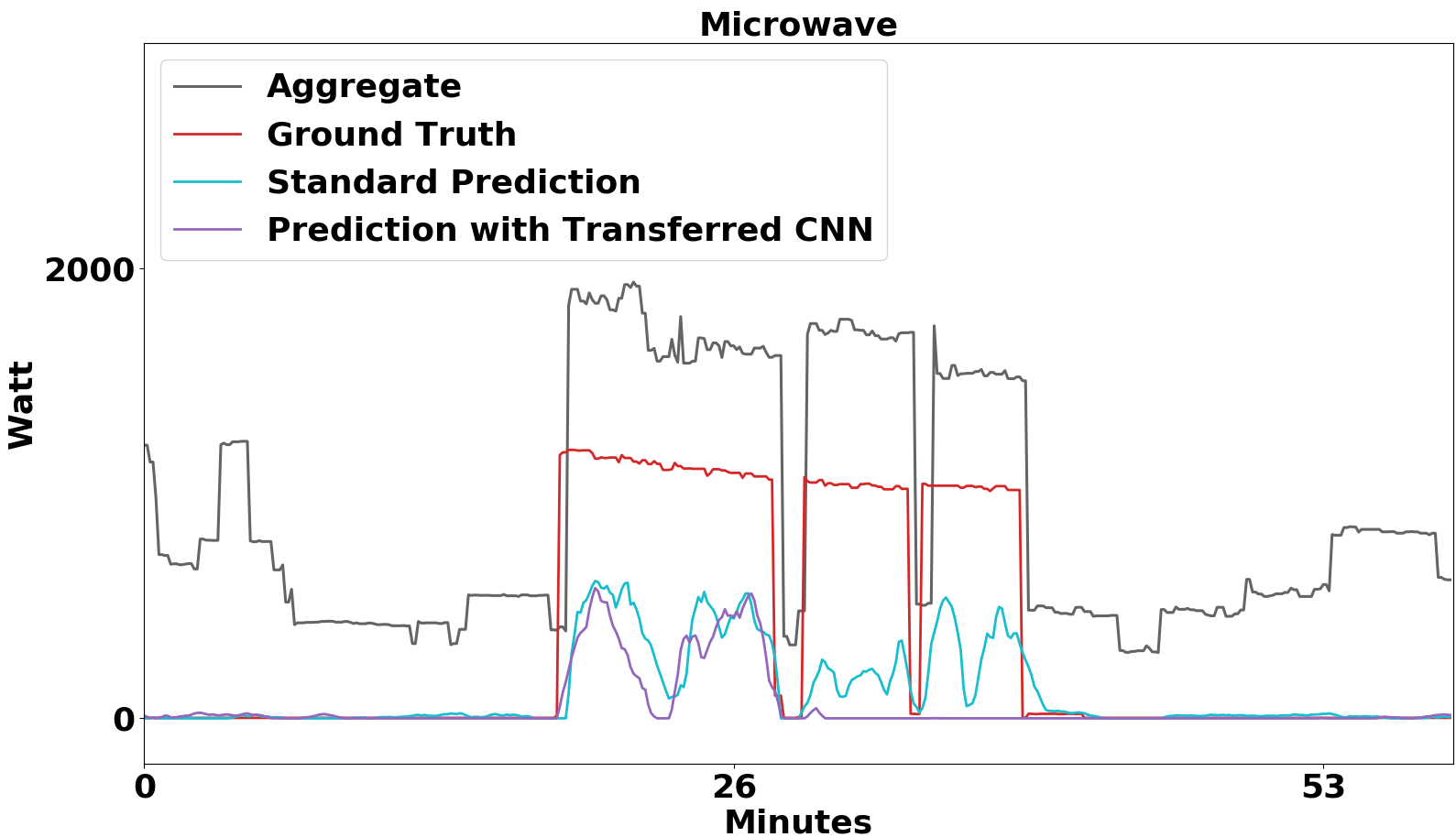}
                \caption{Standard prediction using seq2point model trained on microwave data and the prediction using appliance transfer learning where the CNNs were trained on washing machine data.}
                \label{fig:pred-comp-micro}
            \end{figure}

\begin{figure}
                \centering
                \includegraphics[width=\linewidth,height=1\textheight,keepaspectratio]{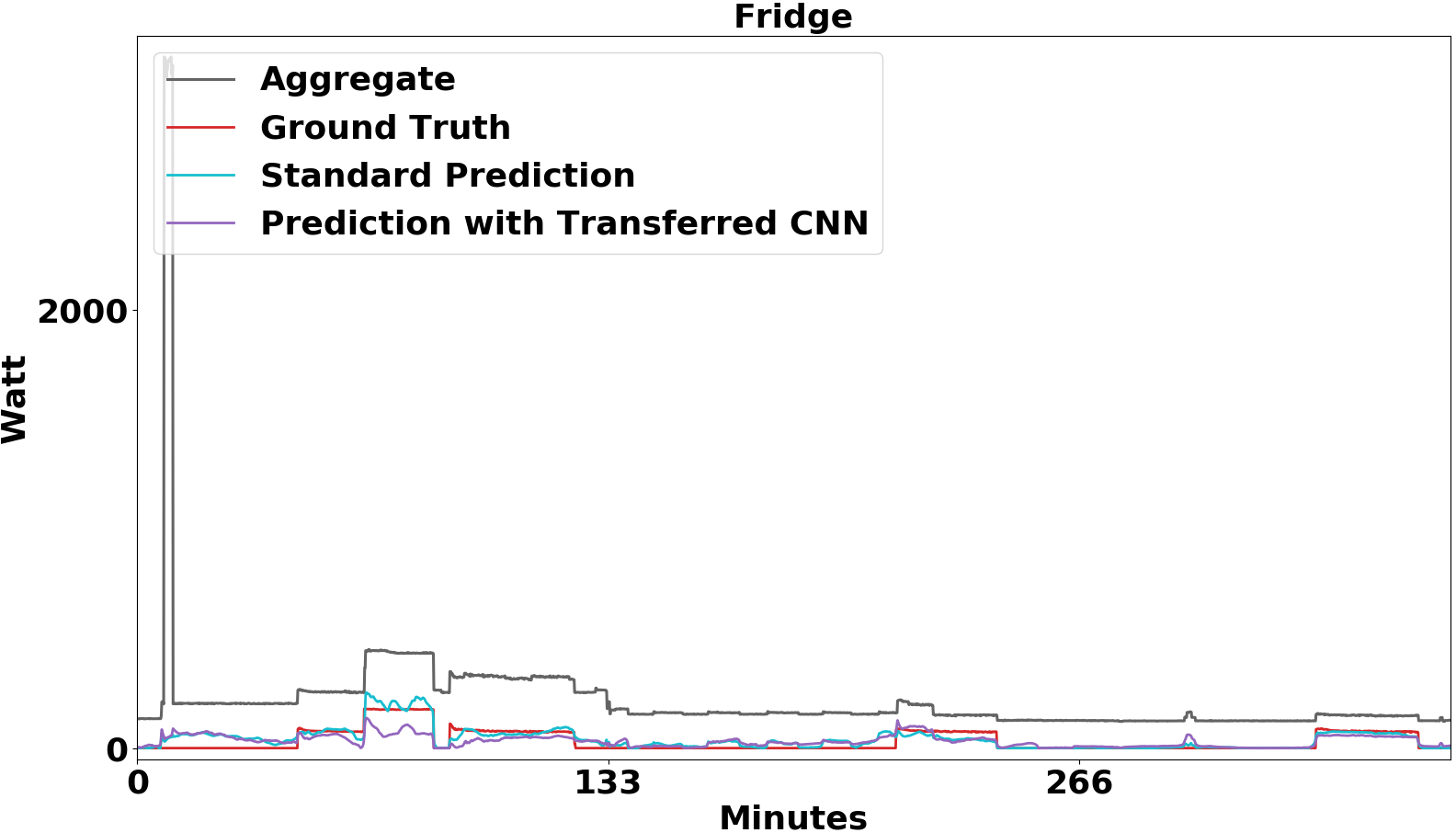}
                \caption{Standard prediction using seq2point model trained on fridge data and the prediction using appliance transfer learning where the CNNs were trained on washing machine data.}
                \label{fig:pred-comp-fridge}
            \end{figure}

\begin{figure}
                \centering
                \includegraphics[width=1\linewidth,height=1\textheight,keepaspectratio]{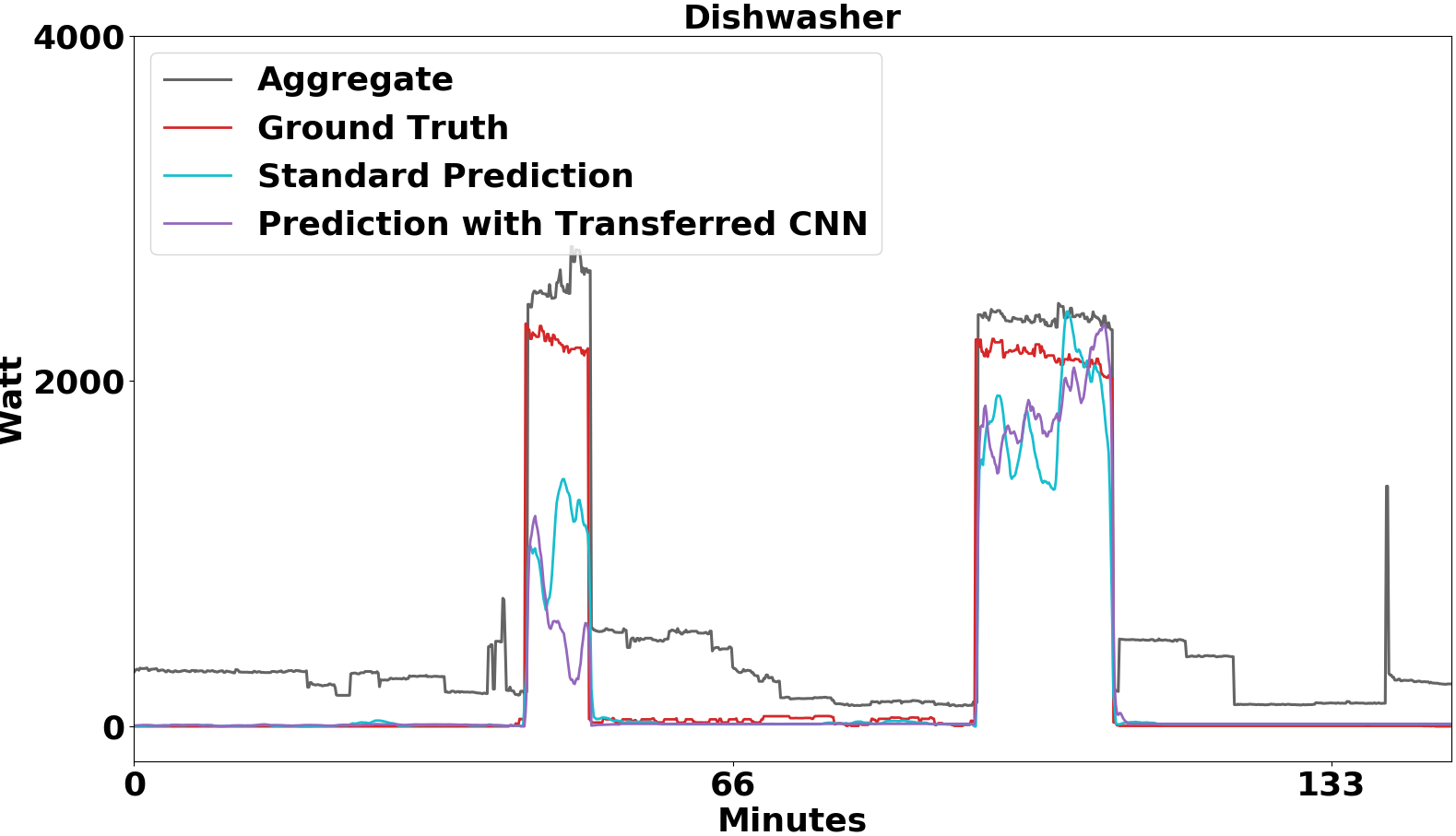}
                \caption{Standard prediction using seq2point model trained on dish washer data and the prediction using appliance transfer learning where the CNNs were trained on washing machine data.}
                \label{fig:pred-comp-dish}
            \end{figure}

    Table \ref{tab:tabrestransF} shows the results of ATL, where the CNN layers were trained on REFIT using washing machine, and tested on REFIT. The results are comparable to the standard training. Figures \ref{fig:wash}, \ref{fig:pred-comp-kettle}, \ref{fig:pred-comp-micro}, \ref{fig:pred-comp-fridge} and \ref{fig:pred-comp-dish} plotted the predicted power reading traces for each appliance, where standard prediction means the prediction produced by the model trained on that appliance, whilst the prediction with transferred CNN means the prediction produced by the ATL. We then examine if the CNN layers trained on REFIT can be transferred to UK-DALE and REDD. Two approaches were examined. Firstly, the CNN layers were trained individually on REFIT for each appliance, and then the fully connected layers were trained on UK-DALE and REDD. The model was then tested on UK-DALE and REDD. The results are shown in the fourth columns in the Table \ref{tab:tabukdale_comp} and \ref{tab:rreddcomp}. Secondly, the CNN layers trained on REFIT were used for all the other appliances. Then the fully connected layers were trained on UK-DALE and REDD. The results are shown in the fifth columns in Table \ref{tab:tabukdale_comp} and \ref{tab:rreddcomp}. 
    
    Comparing the two approaches for transferring CNN models, our results suggest that it is reasonable to use only the CNN layers trained on washing machine for all the appliances. This suggest that we could only need to collect data for washing machine, and thus could reduce the cost for hardware.
        
             \begin{table*}[htbp]
             \caption{Sequence-to-point learning tested on UK-DALE. W.M. denotes Washing Machine.}
                \centering
                \scalebox{1}{
                \begin{tabular}{|l|c|c|c|c|c|c|c|c|c|c|c|c|c|c|c|c|c|c|c|c|}
                \hline
                         &\multicolumn{4}{|c|}{}
                         & \multicolumn{2}{|c|}{Trained on UK-DALE}&  \multicolumn{4}{|c|}{Trained on REFIT}
                         \\                                           
                         Appliance 
                         &\multicolumn{4}{|c|}{AFHMM}
                         &\multicolumn{2}{|c|}{}  &\multicolumn{4}{|c|}{}
                         \\
                         
                        &\multicolumn{4}{|c|}{}
                        &  \multicolumn{2}{|c|}{Tested on UK-DALE \cite{s2p}}&
                        \multicolumn{4}{|c|}{Tested on UK-DALE}
                        \\
                \hline
                                    & \multicolumn{1}{|c|}{MAE} & \multicolumn{1}{|c|}{SAE}&\multicolumn{1}{|c|}{EpD [Wh]}&\multicolumn{1}{|c|}{NDE} &\multicolumn{1}{|c|}{MAE}&\multicolumn{1}{|c|}{SAE}& \multicolumn{1}{|c|}{MAE} & \multicolumn{1}{|c|}{SAE}&\multicolumn{1}{|c|}{EpD [Wh]}&\multicolumn{1}{|c|}{NDE}\\
                \hline
                Kettle  &110.33 &6.94 &1996.82 &5.44        &7.439     &0.069    &6.260    &0.060   &41.08 &0.07 \\
                \hline
                Microwave  &6.50 &0.63 &69.61 &0.99     &8.661     &0.486    &4.770    &0.080  &32.01 &0.26 \\
                \hline
                Fridge   &43.23 &0.99& 896.91 &0.99       &20.894    &0.121    &17.000    &0.090   &113.87 &0.25 \\
                \hline
                Dish w.  &129.71&1.62&127.7&1.41    &27.704    &0.645    &16.490    &0.130   &165.06 &0.18 \\
                \hline
                Washing m. &13.71 &0.95 &316.47 &0.99 &12.663    &0.284   &14.840    &0.500   &229.82 &1.16 \\
                \hline
                Overall  &55.78  &2.51 &944.65  &2.01  &15.472     &0.321    & 11.800    & 0.172   &116.37 &0.384 \\
                mean $\pm$ std   &44.19   &2.37  &709.70 &1.73  &7.718      &0.217   &5.222    &0.166   &74.88 &0.44 \\
                \hline
                \hline
                \end{tabular}
                
                }
                \scalebox{1}{
                \begin{tabular}{|l|c|c|c|c|c|c|c|c|c|c|c|c|c|c|}
                \hline
                         &\multicolumn{4}{|c|}{CNN trained on REFIT, dense}
                         &\multicolumn{4}{|c|}{CNN trained on REFIT (using W.M.),}\\                                           Appliance 
                         &\multicolumn{4}{|c|}{layers trained on UK-DALE}
                         &\multicolumn{4}{|c|}{dense layers trained on UK-DALE}\\
                         
                        &\multicolumn{4}{|c|}{Tested on UK-DALE}
                        &\multicolumn{4}{|c|}{Tested on UK-DALE}\\
                \hline
                                     &\multicolumn{1}{|c|}{MAE}&\multicolumn{1}{|c|}{SAE}& \multicolumn{1}{|c|}{EpD [Wh]}&\multicolumn{1}{|c|}{NDE}& \multicolumn{1}{|c|}{MAE}&\multicolumn{1}{|c|}{SAE}& \multicolumn{1}{|c|}{EpD [Wh]}&\multicolumn{1}{|c|}{NDE}\\
                \hline
                Kettle          &16.879 &0.043 &72.92 & 0.21
                &16.159 & 0.205 &121.03 & 0.35\\
                \hline
                Microwave       &10.973 &0.019 &63.41 &0.81 
                &27.56&0.40&280.27 & 0.72\\
                \hline
                Fridge          &33.078 &0.266 &297.75 &0.51  &
                28.379&0.011&107.09 &0.42\\
                \hline
                Dish w.     &41.106 &0.516 &647.46 &0.74 
                &23.537&0.192&298.89 &0.31 \\
                \hline
                Washing m. &22.941 &0.899 &329.83 &1.19 
                &20.209&0.388&304.11 &0.54 \\
                \hline
                Overall         &24.995 &0.349 &282.27 &0.69 
                &23.169&0.239&222.278 & 0.46\\
                mean $\pm$ std &10.8775 &0.329 &213.35 &0.32 
                &4.57&0.144&88.824 &0.14 \\
                \hline
                \end{tabular}
                }
                \label{tab:tabukdale_comp}
            \end{table*}
    
    \subsubsection{Cross-domain transfer learning}
    In this section, we evaluate seq2point learning on the cross-domain transfer learning (CTL). The CTL here means that the neural networks are trained on a large data, which are then used as the pre-trained model for other data sets, where the neural networks will be fine tuned in the new domain.

    The seq2point models were trained on REFIT. We then tested the trained model on both UK-DALE and REDD. The third column in the Table \ref{tab:tabukdale_comp} shows the CTL results on UK-DALE. The second column in the same table shows the results when the model was trained on REFIT and tested on REFIT. The results show that the performances of the two schemes are similar. This suggests that the model trained on REFIT can be directly applied to UK-DALE for prediction. The second and third columns in Table \ref{tab:rreddcomp} show the results for REDD. However, the performance of seq2point learning was greatly decreased when it was trained on REFIT than when it was trained on REDD. This suggests that the model could not be directly transferred to REDD from REFIT. The intuition is that the signatures of the appliances in REDD are different to those in REFIT, because they are from different countries.
       
   We then consider CTL, which means that the CNN layers are trained on REFIT and then the fully connected dense layers are trained on a small subset taken from REDD and UK-DALE, respectively. The results are shown in the fourth columns in Tables \ref{tab:tabukdale_comp} and \ref{tab:rreddcomp}. It is interesting to note that the performance of seq2point was increased on REDD, but decreased on UK-DALE. It should be noted that the domains of UK-DALE and REFIT are similar. In contrast, the domain of REDD is different to REFIT. This may suggest that the fine tuning on the similar domain may cause overfitting. Fine tuning can help to improve the performance of seq2point learning for different domains.
        
\begin{figure}
                \centering
                \includegraphics[width=\linewidth,height=1\textheight,keepaspectratio]{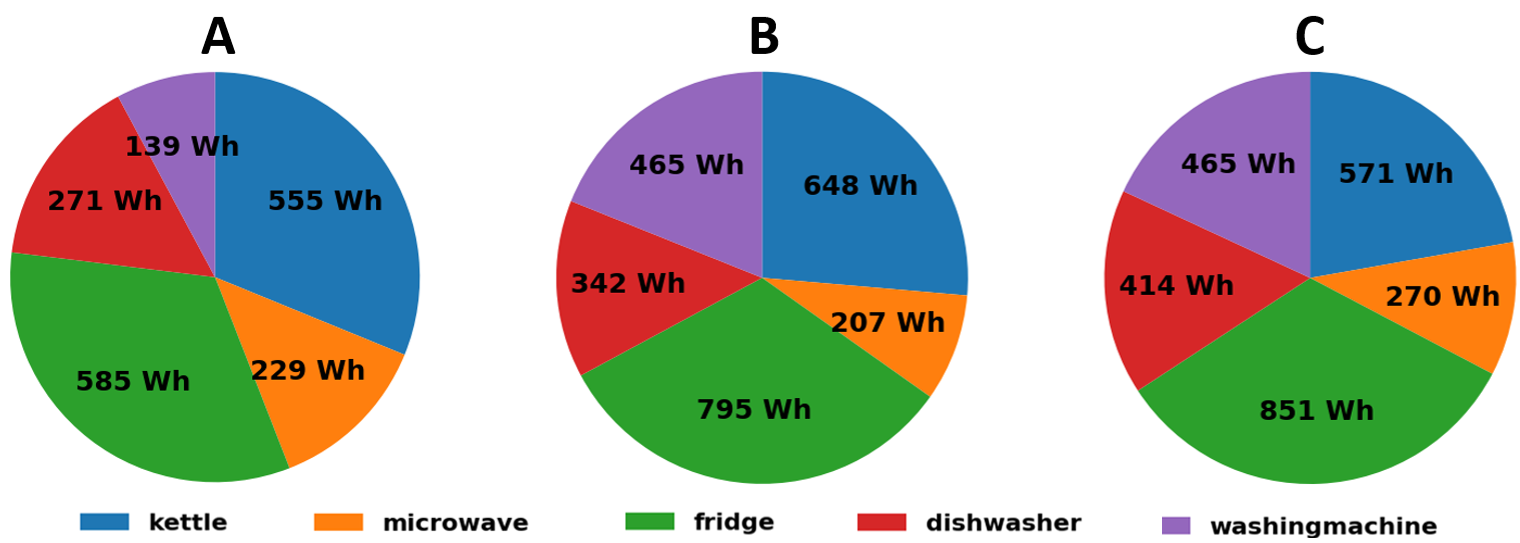}
                \caption{Energy prediction on REFIT using different models. A: the actual energy; B: the model was trained on REFIT; C: the model with frozen CNN which was trained on REFIT using washing machine.}
                \label{fig:refit_piechart}
            \end{figure}

\begin{figure}
                \centering
                \includegraphics[width=\linewidth,height=1\textheight,keepaspectratio]{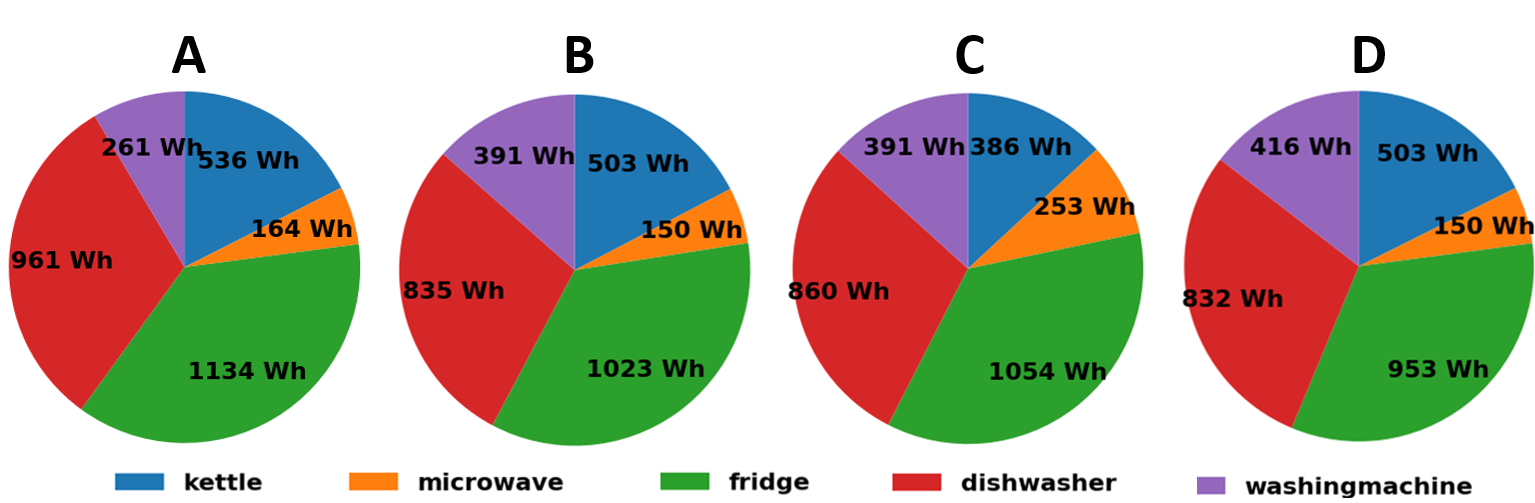}
                \caption{Energy prediction on UK-DALE data using different models. A: the actual energy; B: the model was trained on REFIT; C: the CNN trained on REFIT was forzen and dense layers were trained on UK-DALE; D: the CNN trained on REFIT using washing machine model was frozen and dense layers were trained on UK-DALE.}
                \label{fig:ukdale_piechart}
            \end{figure}

\begin{figure}
                \centering
                \includegraphics[width=\linewidth,height=1\textheight,keepaspectratio]{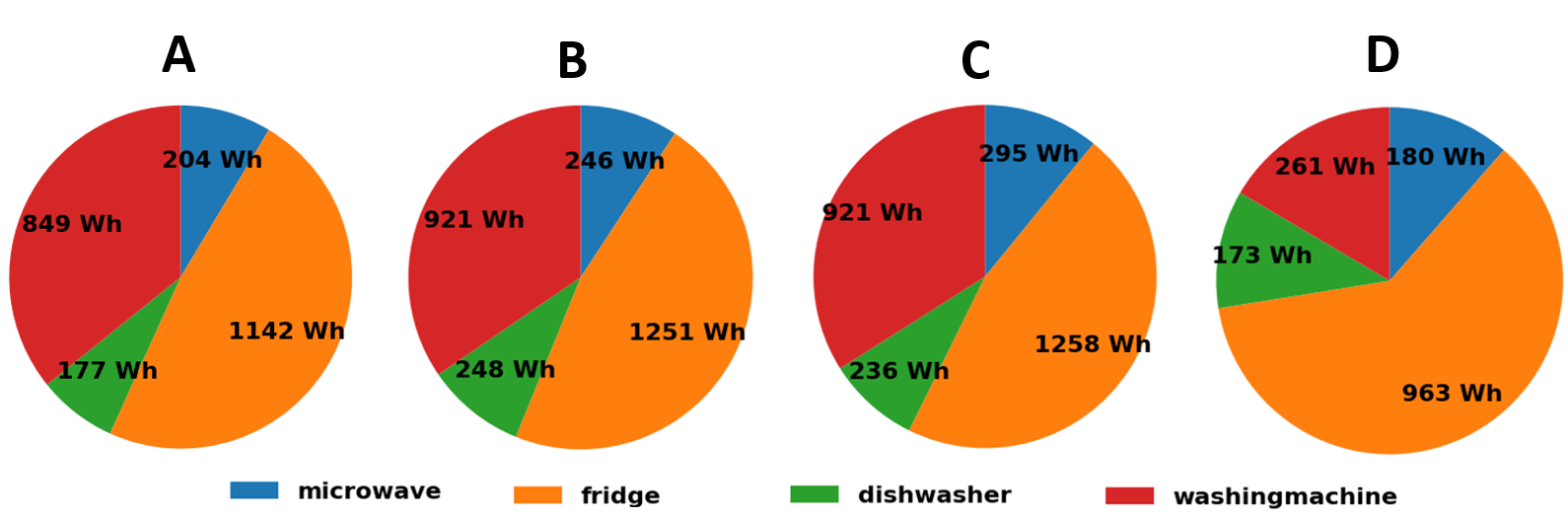}
                \caption{Energy prediction on REDD data using different models. A: the actual energy; B: the model was trained on REFIT; C: the CNN trained on REFIT was forzen and dense layers were trained on REDD; D: the CNN trained on REFIT using washing machine model was frozen and dense layers were trained on REDD.}
                \label{fig:redd_piechart}
            \end{figure}

        We also show the pie charts of energy prediction in Figures \ref{fig:refit_piechart}, \ref{fig:ukdale_piechart} and \ref{fig:redd_piechart} for each data set when the transfer learning methods were applied. It shows that all the transfer learning strategies provide reasonable results for predicting energy consumption of each appliance.
        
        We conclude that when the domains are different, fine tuning does help to improve the performance of seq2point learning. In contrast, when the domains are similar, we may not need fine tuning.
        

             \begin{table*}[htbp]
             \caption{Sequence-to-point learning tested on REDD. W.M. denotes Washing Machine.}
                \centering
                \scalebox{1}{
                \begin{tabular}{|l|c|c|c|c|c|c|c|c|c|c|c|c|c|c|c|c|c|c|c|c|}
                \hline
                         &\multicolumn{4}{|c|}{}
                         & \multicolumn{2}{|c|}{Trained on REDD}&  \multicolumn{4}{|c|}{Trained on REFIT}
                         \\                                           
                         Appliance 
                         &\multicolumn{4}{|c|}{AFHMM}
                         &\multicolumn{2}{|c|}{}  &\multicolumn{4}{|c|}{}
                         \\
                         
                        &\multicolumn{4}{|c|}{}
                        &  \multicolumn{2}{|c|}{Tested on REDD \cite{s2p}}&
                        \multicolumn{4}{|c|}{Tested on REDD}
                        \\
                \hline
                                    & \multicolumn{1}{|c|}{MAE} & \multicolumn{1}{|c|}{SAE}&\multicolumn{1}{|c|}{EpD [Wh]}&\multicolumn{1}{|c|}{NDE} &\multicolumn{1}{|c|}{MAE}&\multicolumn{1}{|c|}{SAE}& \multicolumn{1}{|c|}{MAE} & \multicolumn{1}{|c|}{SAE}&\multicolumn{1}{|c|}{EpD [Wh]}&\multicolumn{1}{|c|}{NDE}\\
                \hline
                Microwave   &11.85   &0.84  &268.63 &0.98  
                            &28.199 &0.059
                            &23.106 &0.357  &208.02 &0.71  \\
                \hline
                Fridge      &69.80 &0.99 &1502.16 &0.99 
                            &28.104 &0.180  
                            &38.637 &0.022  &205.48 &0.36 \\
                \hline
                Dish w.     &155.25 &7.19 &3427.86 &9.36    
                            &20.048 &0.567
                            &29.677 &0.711  &499.52 &1.02\\
                \hline
                Washing m.  &14.25 &0.07 &84.53 &84.53 
                            &18.423 &0.277
                            &36.832 &0.736  &750.85 &0.91 \\ 
                \hline
                Overall     &62.79 &2.28 &1320.80 &2.88
                            &23.693 &0.270
                            &32.063 &0.457  &415.98&0.75 \\ 
                mean $\pm$ std  
                            &58.20 &2.86 &1333.04 &3.76
                            &4.494 &0.187
                            &6.162  &0.292  &227.29 &0.25 \\
                \hline
                \hline
                \end{tabular}
                
                }
                \scalebox{1}{
                \begin{tabular}{|l|c|c|c|c|c|c|c|c|c|c|c|c|c|c|}
                \hline
                         &\multicolumn{4}{|c|}{CNN trained on REFIT, dense}
                         &\multicolumn{4}{|c|}{CNN trained on REFIT (using W.M.),}\\                                           Appliance 
                         &\multicolumn{4}{|c|}{layers trained on REDD}
                         &\multicolumn{4}{|c|}{dense layers trained on REDD}\\
                         
                        &\multicolumn{4}{|c|}{Tested on REDD}
                        &\multicolumn{4}{|c|}{Tested on REDD}\\
                \hline
                                     &\multicolumn{1}{|c|}{MAE}&\multicolumn{1}{|c|}{SAE}& \multicolumn{1}{|c|}{EpD [Wh]}&\multicolumn{1}{|c|}{NDE}& \multicolumn{1}{|c|}{MAE}&\multicolumn{1}{|c|}{SAE}& \multicolumn{1}{|c|}{EpD [Wh]}&\multicolumn{1}{|c|}{NDE}\\

                \hline
                Microwave       &27.792 &0.023 &247.57 &0.68 
                                &13.806 &0.144 &80.90  &1.00 \\
                \hline
                Fridge          &34.906 &0.057 &118.02 &0.29 
                                &35.932 &0.285 &303.24 &0.54 \\
                \hline
                Dish w.     &25.002 &0.007 &274.62 &0.35 
                            &13.597&0.122&234.85 &1.79 \\
                \hline
                Washing m. &17.991 &0.128 &181.31 &0.19 
                            &43.775 &0.692 &826.08 &0.96 \\
                \hline
                Overall         &26.422 &0.054 &205.38 &0.37
                                &26.778&0.311&361.27   &1.07\\
                mean $\pm$ std &6.061 &0.047 &60.80 &0.18
                                &13.367&0.229&280.18 &0.45\\
                \hline
                \end{tabular}
                }
                \label{tab:rreddcomp}
            \end{table*}

\section{Conclusions}\label{Conclusion}
This work presents transfer learning for NILM. Typically, we considered the transferability of seq2point learning. In the previous studies, seq2point learning and testing were only considered in the same domain. In contrast, we consider to learn the model in one domain and test in a different domain in this paper. Our hypothesis was that the appliance features extracted by using convolutional neural networks are invariant across appliances and as well as across data domains. Our experiments on transfer learning support our hypothesis. Two different transfer learning approaches are studied experimentally, which are appliance transfer learning and cross-domain transfer learning. In our vision, by means of transfer deep learning, it could be possible to find a unique model for residential appliance worldwide that can be then adapted by using reduced training data, thus achieving the oracle NILM, remarkable computational savings and as well as a decreased dependency on specific appliance-labelled data.

Seq2point models were trained on REFIT, and then tested on REDD and UK-DALE. Our conclusions are the following. Firstly, concerning ATL, the CNN layers trained on washing machines can be applied to other appliances. Basically, this indicates that all appliances use the similar signatures for NILM purposes. Therefore, we can use a large data to train the CNN layers, and then apply the trained CNN layers to any other unseen data to obtain the signatures. The fully connect layers can then be trained solely on the unseen data. Secondly, regarding CTL, we found that if the domains of training and testing data are similar, the trained model may not require fine tuning; if the domains are different, fine tuning does help to improve the performance of the trained model. The intuition is that if the domains are similar, fine tuning may lead to over-fitting since only a small subset was used for fine tuning. On the other hand, if the domains are different, fine tuning does help the model to adapt for the unseen domain.

The benefits introduced by the adoption of a transfer learning strategy are: Firstly and most importantly, it provides a possible approach to achieve the oracle NILM; precisely, it could provide a unique model for all residential appliances worldwide; Secondly, it could potentially reduce the number of sensors for each appliance to be installed in households since the trained features could be transferred to other appliances or domains and thus reduce financial cost; Thirdly, transfer learning does offer remarkable computational savings since pre-trained models can be reused for other appliance or domains.

\ifCLASSOPTIONcaptionsoff
  \newpage
\fi



\bibliographystyle{IEEEtran}
\bibliography{references.bib}
%


%

\begin{IEEEbiography}[{\includegraphics[trim={12cm 9cm 32cm 7cm},width=1in,height=1.25in,clip]{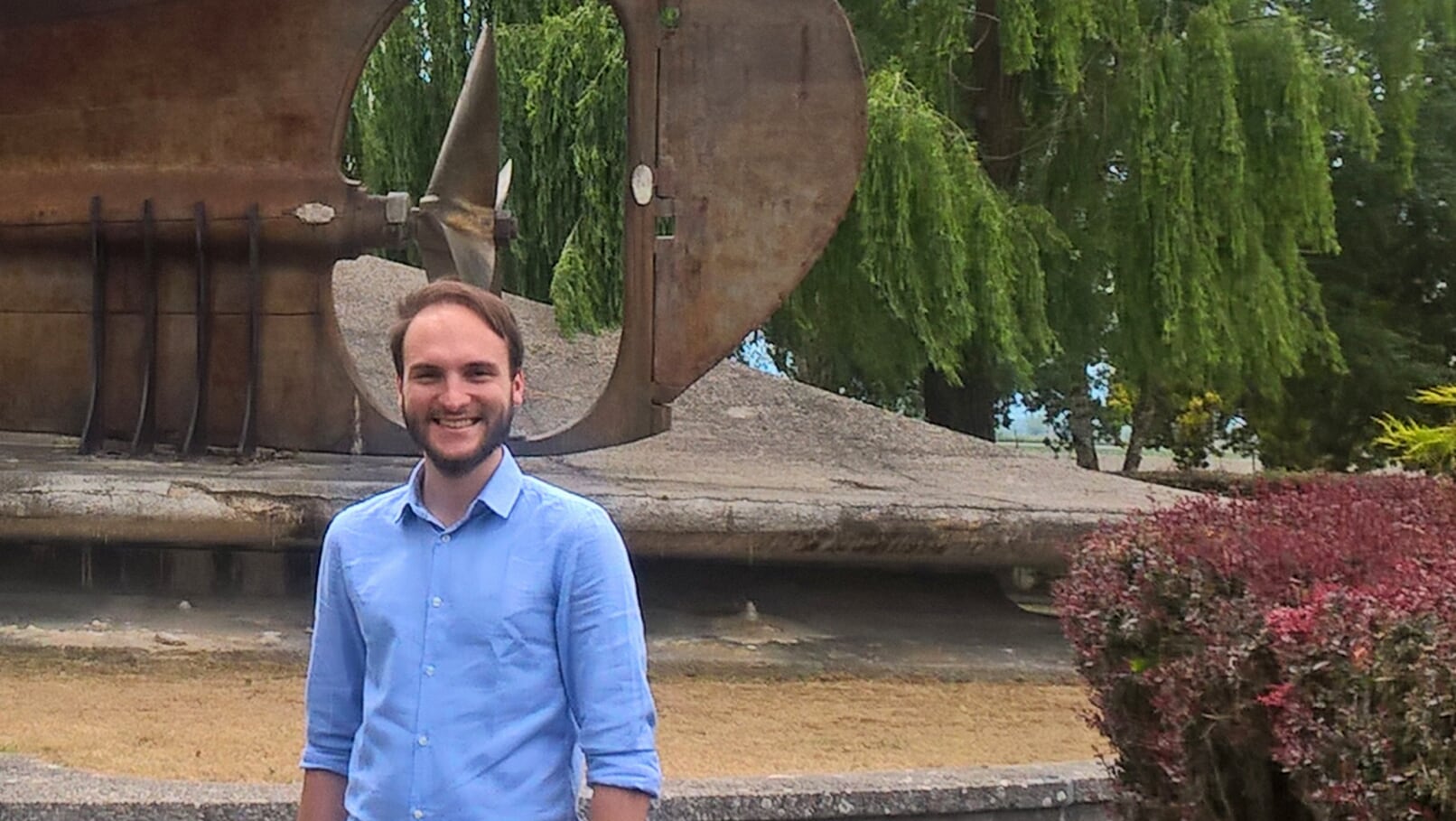}}]{Michele D'Incecco} was born in Recanati, Italy on July 1992. He received the Bachelor's degree in Electronics and Telecommunications Engineering in 2012 Polytechnic University of Marche, (UnivPM), Ancona. He recently got a MS in Electronics Engineering in 2018 from UnivPM. His main interests focused on energy management and deep learning techniques applied to energy disaggregation. 
\end{IEEEbiography}


\begin{IEEEbiography}[{\includegraphics[width=1in,height=1.25in,clip,keepaspectratio]{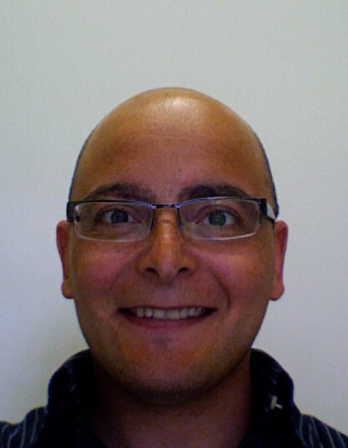}}]{Stefano Squartini} (IEEE Senior Member, IEEE CIS Member) was born in Ancona, Italy, on March 1976. He got the Italian Laurea with honors in electronic engineering from University of Ancona (now Polytechnic University of Marche, UnivPM), Italy, in 2002. He obtained his PhD at the same university (November 2005). He worked also as post-doctoral researcher at UnivPM from June 2006 to November 2007, when he joined the DII (Department of Information Engineering) as Assistant Professor in Circuit Theory. He is now Associate Professor at UnivPM since November 2014. His current research interests are in the area of computational intelligence and digital signal processing, with special focus on speech/audio/music processing and energy management. He is author and coauthor of more than 200 international scientific peer-reviewed articles. He is Associate Editor of the IEEE Transactions on Neural Networks and Learning Systems, IEEE Transactions on Cybernetics and IEEE Transactions on Emerging Topics in Computational Intelligence, and also member of Cognitive Computation, Big Data Analytics and Artificial Intelligence Reviews Editorial Boards. He joined the Organizing and the Technical Program Committees of more than 70 International Conferences and Workshops in the recent past. He is the Chair of the IEEE CIS Task Force on Computational Audio Processing and the leading organiser of the International Workshop on Computational Energy Management in Smart Grids.
\end{IEEEbiography}



\begin{IEEEbiography}[{\includegraphics[width=1in,height=1.25in,clip,keepaspectratio]{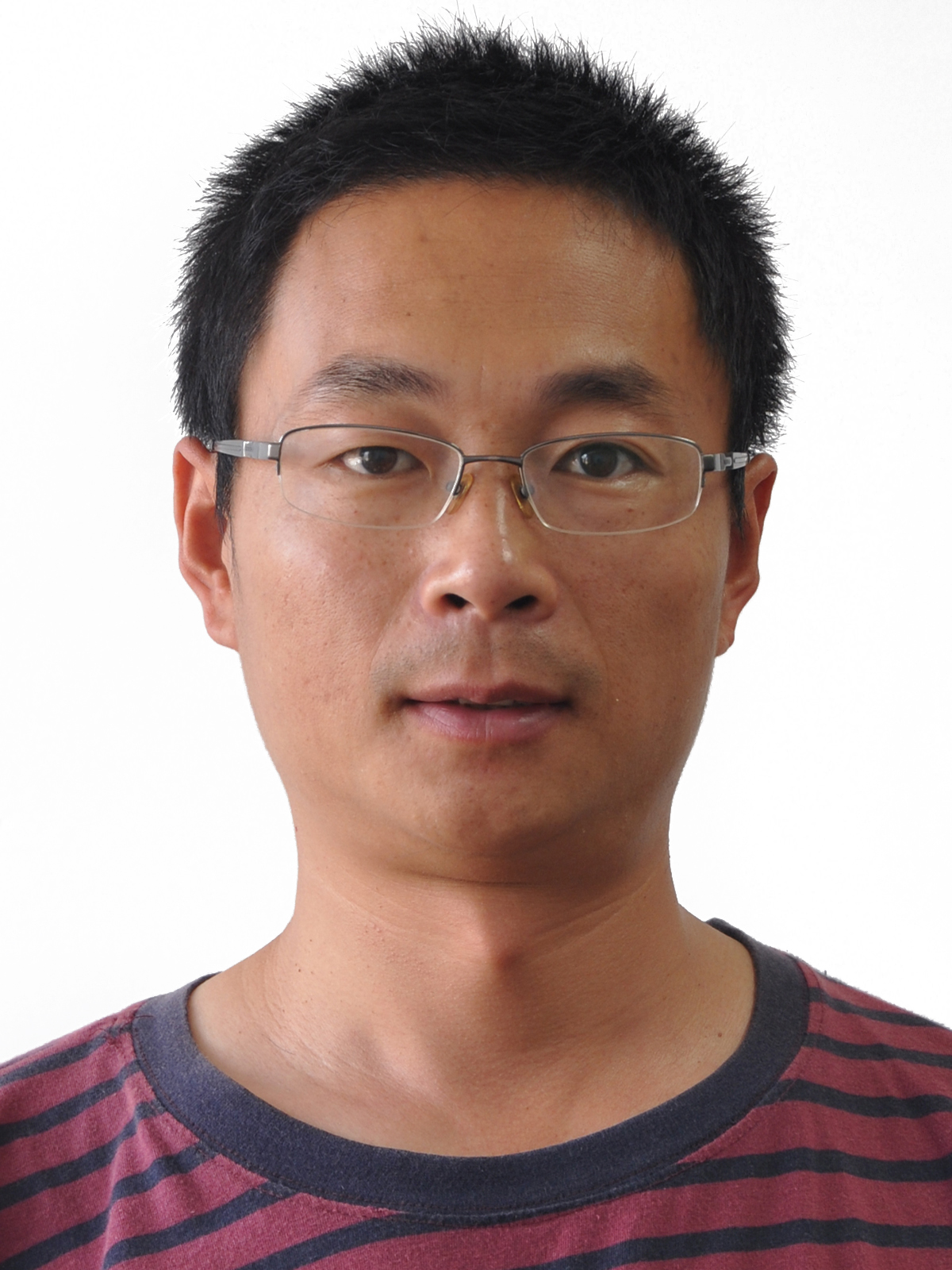}}]{Mingjun Zhong} was born in Wei-Fang, Shan-Dong province, China. He obtained his BSc (July 1999) and PhD (July 2004) both in Applied Mathematics at the Dalian University of Technology, China. He was an associate professor at the Dalian Nationalities University, China. He worked as a post-doctoral researcher at INRIA Rennes in France from December 2015 to December 2016. He worked as a Research Assistant in the School of Computing at the University of Glasgow in UK from February 2008 to October 2011. He also worked as a Research Associate in the School of Informatics at the University of Edinburgh in UK from July 2013 to April 2017. He was an Associate Professor at the Dalian University of Technology. He is now a Senior Lecture at the University of Lincoln in UK. His research interests are Machinle Learning, Bayesian Inference, and Deep Learning. 
\end{IEEEbiography}



\end{document}